\documentclass[graybox]{svmult}

\usepackage{type1cm}        
\usepackage{makeidx}         
\usepackage{graphicx}        
\usepackage{multicol}        
\usepackage[bottom]{footmisc}

\usepackage{newtxtext}       %
\usepackage[varvw]{newtxmath}       

\usepackage{algorithm}
\usepackage{algorithmic}

\usepackage{newfloat}
\usepackage{listings}
\floatstyle{ruled}
\newfloat{listing}{tb}{lst}{}
\floatname{listing}{Listing}

\usepackage{changepage}
\usepackage{tcolorbox}
\tcbuselibrary{most}
\usepackage{multirow}
\usepackage{booktabs}
\usepackage[textwidth=\marginparwidth, textsize=tiny, color=green!30]{todonotes}
\usepackage{pifont}
\newcommand{\cmark}{\ding{51}}%
\newcommand{\xmark}{\ding{55}}%

\usepackage{titlesec}
\titlespacing*{\section}{0pt}{5pt}{5pt}
\titlespacing*{\subsection}{0pt}{5pt}{5pt}
\titlespacing*{\subsubsection}{0pt}{5pt}{5pt}

\usepackage[sort&compress,numbers]{natbib}

\definecolor{veryLightGray}{RGB}{230, 230, 230}
\definecolor{darkerGray}{RGB}{180,180,180}
\definecolor{lightBlue}{RGB}{0, 153, 255}
\definecolor{lightRed}{RGB}{204, 0, 255}

\lstdefinelanguage{Solidity}{
	keywords=[1]{anonymous, assembly, assert, balance, break, call, callcode, case, catch, class, constant, continue, constructor, contract, debugger, default, delegatecall, delete, do, else, emit, event, experimental, export, external, false, finally, for, function, gas, if, implements, import, in, indexed, instanceof, interface, internal, is, length, library, log0, log1, log2, log3, log4, memory, modifier, new, payable, pragma, private, protected, public, pure, push, require, return, returns, revert, selfdestruct, send, solidity, storage, struct, suicide, super, switch, then, this, throw, transfer, true, try, typeof, using, value, view, while, with, addmod, ecrecover, keccak256, mulmod, ripemd160, sha256, sha3}, 
	keywordstyle=[1]\color{blue}\bfseries,
	keywords=[2]{address, bool, byte, bytes, bytes1, bytes2, bytes3, bytes4, bytes5, bytes6, bytes7, bytes8, bytes9, bytes10, bytes11, bytes12, bytes13, bytes14, bytes15, bytes16, bytes17, bytes18, bytes19, bytes20, bytes21, bytes22, bytes23, bytes24, bytes25, bytes26, bytes27, bytes28, bytes29, bytes30, bytes31, bytes32, enum, int, int8, int16, int24, int32, int40, int48, int56, int64, int72, int80, int88, int96, int104, int112, int120, int128, int136, int144, int152, int160, int168, int176, int184, int192, int200, int208, int216, int224, int232, int240, int248, int256, mapping, string, uint, uint8, uint16, uint24, uint32, uint40, uint48, uint56, uint64, uint72, uint80, uint88, uint96, uint104, uint112, uint120, uint128, uint136, uint144, uint152, uint160, uint168, uint176, uint184, uint192, uint200, uint208, uint216, uint224, uint232, uint240, uint248, uint256, var, void, ether, finney, szabo, wei, days, hours, minutes, seconds, weeks, years},	
	keywordstyle=[2]\color{teal}\bfseries,
	keywords=[3]{block, blockhash, coinbase, difficulty, gaslimit, number, timestamp, msg, data, gas, sender, sig, value, now, tx, gasprice, origin},	
	keywordstyle=[3]\color{violet}\bfseries,
	identifierstyle=\color{black},
	sensitive=true,
	comment=[l]{//},
	morecomment=[s]{/*}{*/},
	commentstyle=\color{gray}\ttfamily,
	stringstyle=\color{red}\ttfamily,
	morestring=[b]',
	morestring=[b]"
}

\lstdefinelanguage{N3}
{
    sensitive = true,
    keywords = [1]{Class, EquivalentTo, SubClassOf},
    morekeywords = [3]{rdf:type, math:exponentiation, math:quotient},
    morekeywords = [4]{exam, value, patient, profile, sys, sysValue, dias, diasValue, plan, ethnicity, demo, NewTreatmentSubPlan, RecommendDiabetesScreening, h, w, h_exp, bmi},
    keywordstyle=[3]\color{lightRed}\textbf,
    keywordstyle=[4]\color{lightBlue}\textbf,
    morestring=[b]'',
    alsoletter=:
}

\newtcolorbox[auto counter]{myexample}[2][]{
  enhanced,
  breakable,
  left=2pt,
  right=2pt,
  top=1pt,
  bottom=1pt,
  fonttitle=\scshape,
  title=Example~\thetcbcounter: #2,
  #1,
}
\newtcolorbox[auto counter]{scenario}[1][]{
  enhanced,
  breakable,
  left=2pt,
  right=2pt,
  top=1pt,
  bottom=1pt,
  fonttitle=\scshape,
  title=Scenario~\thetcbcounter,
  #1,
}

\makeindex             

\usepackage{wrapfig}

\begin{document}

\newcommand{\backticks}{\`{}\`{}\`{}}

\title*{Using Large Language Models for Generating Smart Contracts for Health Insurance from Textual Policies}
\author{Inwon Kang\orcidID{0000-0001-8912-287X} and\\William Van Woensel\orcidID{0000-0002-7049-8735} and\\ Oshani Seneviratne\orcidID{0000-0001-8518-917X}}
\institute{Inwon Kang \at Rensselaer Polytechnic Institute, Troy, New York, USA \email{kangi@rpi.edu}
\and William Van Woensel \at University of Ottawa, Ottawa, Ontario, Canada \email{wvanwoen@uottawa.ca}
\and Oshani Seneviratne \at Rensselaer Polytechnic Institute, Troy, New York, USA \email{senevo@rpi.edu} }
%
%
\titlerunning{LLMs for Health Insurance Policies}
\authorrunning{Kang, Van Woensel, and Seneviratne}

\maketitle

\abstract{We explore using Large Language Models (LLMs) to generate application code that automates health insurance processes from text-based policies. 
We target blockchain-based smart contracts as they offer immutability, verifiability, scalability, and a trustless setting: any number of parties can use the smart contracts, and they need not have previously established trust relationships with each other.
Our methodology generates outputs at increasing levels of technical detail: (1) textual summaries, (2) declarative decision logic, and (3) smart contract code with unit tests.
We ascertain LLMs are good at task (1), and the structured output is useful to validate tasks (2) and (3). Declarative languages (task 2) are often used to formalize healthcare policies, but their execution on blockchain is non-trivial.
Hence, task (3) attempts to directly automate the process using smart contracts.
To assess the LLM output, we propose \emph{completeness}, \emph{soundness}, \emph{clarity}, \emph{syntax}, and \emph{functioning code} as metrics.
Our evaluation employs three health insurance policies (\textit{scenarios}) with increasing difficulty from Medicare's official booklet. 
Our evaluation uses GPT-3.5 Turbo, GPT-3.5 Turbo 16K, GPT-4, GPT-4 Turbo and CodeLLaMA.
Our findings confirm that LLMs perform quite well in generating textual summaries.
Although outputs from tasks (2)-(3) are useful starting points, they require human oversight:
in multiple cases, even ``runnable'' code will not yield sound results; the popularity of the target language affects the output quality; and more complex scenarios still seem a bridge too far.
Nevertheless, our experiments demonstrate the promise of LLMs for translating textual process descriptions into smart contracts. }

\section{Introduction}
Many components of healthcare still rely on manual processes governed by textual policies and guidelines.
For example, Centers for Medicare \& Medicaid Services (CMS) utilizes internet- and paper-based manuals to administer their programs~\cite{cms-manuals}. 
In health insurance, an insurance company with a written policy will require a human operator to go back and forth to ensure reimbursement criteria are met to finalize the payment, possibly causing payment delays and reimbursement mistakes. 
Therefore, auto-generating code can save many hours of human effort and reduce errors.
Such automation allows for real-time, error-free execution of health insurance policies, reducing administrative overhead and the potential for human error. 

Smart contracts on blockchain can be used to automate healthcare insurance processes. In general, blockchain offers \emph{trustlessness}, \emph{immutability} and \emph{verifiability}: no trust is required in a single party, but rather the underlying consensus mechanism to validate the transactions, and all involved parties can verify their transactions in an immutable ledger.
Smart contracts offer an extension of these properties: 
after publishing on the blockchain, a smart contract cannot be modified, its execution cannot be tampered with, and each execution is similarly verifiable.
Health insurance processes, among others (such as clinical decision support systems~\cite{sc_gen_2023}), are excellently suited to leverage these properties, as the parties involved---patients, care providers, and insurers---have competing interests and often a mutual distrust, respectively benefiting from lowest-cost options and maximizing reimbursement and health benefits.
Indeed, smart contracts offer an immutable record of transactions and rules that enhance transparency in health insurance policy execution, which is crucial for building trust between policyholders and insurers, particularly in scenarios involving claim disputes.
For instance, blockchain technology has been used to secure access to medical data for insurance companies, hospitals, and doctors~\cite{ben2020application}, 
and for preference-based selection of suitable insurance policies~\cite{Zhou2018MIStoreAB}.



That said, automating interactions between healthcare parties, albeit on blockchain or another platform, is still a complex problem.
Implementing a described process from dense natural text, such as ``legalese'' insurance policies, is error-prone and labor-intensive; 
requiring domain expertise which a software/Web3 (smart contract) developer typically lacks.
Indeed, decision logic in healthcare is typically designed by domain experts, using a general-purpose (e.g., N3~\cite{notation3_2023} or domain-specific formalism  (e.g., PROforma~\cite{sutton_syntax_2003}, SDA~\cite{Riano07}, GLEAN~\cite{van_woensel_explainable_2022}).
Recently, Large Language Models (LLM), such as (but not limited to) OpenAI's ChatGPT and GPT-4~\cite{openaiGPT4TechnicalReport2023}
and Meta's LLaMA-2~\cite{touvronLLaMAOpenEfficient2023} 
have shown impressive advancements in automating tasks such as summarizing a document or writing code.
If an LLM can automate some part of translating dense natural text into decision logic or smart contracts, this process will become less labor-intensive. 
Hence, by using LLMs to generate smart contract code for health insurance policies, we can further streamline policy enforcement: smart contracts can be dynamically generated by LLMs in case of changing regulations, ensuring that health policy processes remain compliant over time without manual intervention.

We study the effectiveness of LLMs in translating health insurance policies from textual sources into smart contracts using a workflow.
We present an initial methodology featuring 3 tasks: 
(1) textual summaries, which are useful to validate following steps compared to natural text; 
(2) declarative decision logic, often used to formalize healthcare policies but difficult to deploy on blockchain~\cite{sc_gen_2023}; and 
(3) smart contract code with unit tests. 

\section{Methodology}

In developing our methodology, we carefully engineered a single prompt for each task based on established best practices for prompt engineering with LLMs. The quality of the prompt is fundamental to the success of the task, and as such, we ensured that each prompt was clear, unambiguous, specific to the task, and concise.

\subsection{Code Generation Tasks}
\begin{figure}
    \centering
    \includegraphics[width=\linewidth]{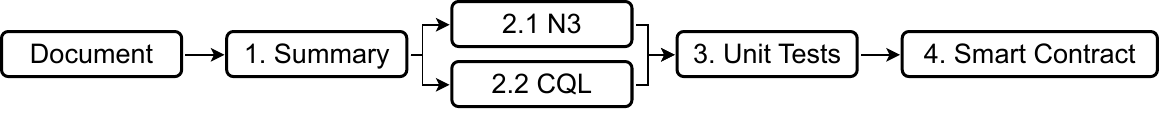}
    \caption{Combination of individual tasks}
    \label{fig:task-combinations}
\end{figure}

We consider the code generation tasks described in the sections below as illustrated in Fig. \ref{fig:task-combinations}.

\subsubsection{Task 1: Writing a structured text summary.} 
The LLM is asked to provide a concise summary of the natural text, as well as a bulleted list of decisional criteria described therein.
A natural language text may contain unnecessary repetitions and be difficult to follow, especially when written using a legal or medical style. 
A structured summary could help a domain expert or Web3 developer identify core concepts and decision logic to validate subsequent steps better.

\subsubsection{Task 2: Extracting high-level decision logic. }
The LLM is asked to extract high-level decision logic using multiple formalisms, which encodes the process and decisional criteria found in the natural text.
This decision logic unambiguously formalizes the process using a given formalism; the resulting model can be executed on multiple platforms using an execution engine. 
The deployment of N3 models has been explored on blockchain~\cite{sc_gen_2023} but this work is still preliminary.
We consider the following LLM generation targets for decision logic:
\begin{enumerate}
    \item \textit{Notation3 (N3)}
    ~\cite{notation3_2023}, is an expressive, human-readable, and domain-agnostic rule language that can be used to capture complex decision logic.
    It extends the Resource Description Framework (RDF) with rules, graph terms, built-ins, and (scoped) negation as failure. N3's expressiveness, with its ability to describe rules and logic through a declarative human-readable form, makes it well-suited for articulating the high-level decision-making processes in health insurance claims.
    \item \textit{Clinical Quality Language (CQL)}
    ~\cite{cql_hl7}, is a high-level, domain-specific formalism designed to capture clinical quality and decision-support logic.
    It is part of an international standard for health data interoperability, Fast Healthcare Interoperability Resources (FHIR)~\cite{hl7_fhir}. Given that health insurance policies are inherently tied to clinical outcomes and healthcare delivery, CQL provides a natural and direct means of encoding rules related to health insurance policies.
\end{enumerate}

\subsubsection{Task 3: Generating unit tests.}
The LLM is asked to generate unit tests for a smart contract. This can help verify the correctness of the smart contract code and help the human programmer gain a better understanding of how the final code should behave.
We choose to use Remix IDE's testing plugin~\cite{remix_ide_unit_testing}.
Our decision is motivated by the fact when prompted for a solidity contract unit test, many LLMs generate code that uses Remix's testing plugin, which suggests that Remix, a browser-based IDE and a popular choice among many Web3 developers, is probably the most familiar method for LLMs for smart contract testing.

\subsubsection{Task 4: Generating smart contract code.}
The LLM is asked to generate smart contract code that automates processes found in natural text on a blockchain.
As the programming language, we target \textit{Solidity}~\cite{solidity}, originally designed by the Ethereum foundation.
It has since gained widespread popularity: it is supported by all Ethereum-compatible blockchains such as Polygon and Avalanche, and even private permissioned blockchains such as Hyperledger Fabric~\cite{hyperledger}.
Hence, a Solidity smart contract can be used in a range of blockchains.


\subsection{Prompt Engineering}
\label{sec:prompt_engineering}

Previous works using LLMs have found that the effectiveness of LLM depends on how LLM queries---or prompts---are phrased.
We engineered our prompts following best practices guides provided by OpenAI~\cite{openai_2023_best_practices} and Deep learning AI~\cite{deeplearning_ai_chatgpt_prompt_eng,whitePromptPatternCatalog2023}.
These resources suggest providing the \textit{role} of the LLM and \textit{rules/instructions} to be followed alongside the \textit{input}.
We similarly break down the prompt into different parts, each of which playing a distinct role.


{
\begin{myexample}[label=block:prompt_example]{Prompt for task 1}
    \textbf{System:}
    \begin{adjustwidth}{0.5cm}{}
1.      You are a healthcare expert who translates \{INPUT\_NAME\} into a more comprehensible format for a Web3 developer. 

2.      Your task is to provide a high-level summary given in the \{INPUT\_NAME\}.

3.      The requirements in the \{INPUT\_NAME\} must be summarized in a numbered list format. If a requirement refers to another requirement which was mentioned previously, refer to it using the numbering in the list. You must capture every single requirement described in the \{INPUT\_NAME\}.\\
        
    \end{adjustwidth}

    \textbf{User:} \{INPUT\_DOCUMENT\}
\end{myexample}
}

Example~\ref{block:prompt_example} shows an example prompt for task 1 (structured text summary), where \{INPUT\_NAME\} and \{INPUT\_DOCUMENT\} are placeholders. We outline the different parts below:
\begin{enumerate}
  \item \textbf{Role}: One sentence description of the role the LLM is expected to play in the conversation.
  \item \textbf{Task Description}: One sentence description of the task.
  \item \textbf{Task Detail}: Multi-sentence description of the rules. 
\end{enumerate}


\section{Experiment Setup}


We evaluate our methodology on multiple health insurance scenarios, using four variants of OpenAI's GPT models (GPT 3.5 Turbo, GPT 3.5 Turbo-16K, GPT 4 and GPT 4 Turbo)~\cite{openai_2023_models} and Meta's CodeLLaMA-instruct-34B~\cite{roziereCodeLlamaOpen2023,codellama_2023_codelama34b}. 
We chose these models because they were the most recent and accessible LLMs available at the time of writing this paper, i.e., mid-November 2023. 
The OpenAI models were accessed through the official API, and CodeLLaMA was accessed through a locally hosted API~\cite{oobabooga_2023_text_generation_webui} on a machine equipped with one A100 (80GB VRAM) system. 
For reproducibility, we set the temperature of the models to 0 for OpenAI and near-zero value for CodeLLaMA. The temperature value decides randomness and is added to the next word picked by the LLM. A low temperature means that the LLM will likely generate the same response to the same input.

The three scenarios ultimately chosen were based on their increasing level of complexity. The scenarios also reflect the most common situations encountered in health insurance claims and policy enforcement.

\subsection{Experiment Scenarios}

In selecting the scenarios for our experiment, we employed a multi-step approach to ensure that they were representative of a broad range of health insurance policies. 
First, we reviewed existing health insurance policies from multiple countries to identify common features and requirements that are applicable across different healthcare systems.
We then analyzed a diverse set of US-based health insurance policies obtained from public databases and insurance providers that cover a spectrum of policy types, including general health coverage, specialized treatments, and chronic disease management.
Each scenario outlined below represents different aspects of health insurance policies, such as preventative care incentives, coverage determination for different treatment types, and claims adjudication processes.

\begin{scenario}[label=block:summary_scenario1]
Medicare will cover injectable osteoporosis drugs if the patient is a woman meeting the criteria for the home health benefit, suffering from a bone fracture due to postmenopausal osteoporosis, and cannot self-inject.
Medicare may cover a home health nurse to provide the injection.
\end{scenario}

\begin{scenario}[label=block:summary_scenario2]
Medicare will cover home-based Intravenous Immune Globulin (IVIG) if the patient is diagnosed with primary immunodeficiency disease and it is deemed medically appropriate to get the IVIG at home.
\end{scenario}

\begin{scenario}[label=block:summary_scenario3]
Medicare will cover transplant drug therapy if Medicare Part A helps to pay for the organ transplant and the patient has Medicare Part B at the time of drug prescription.
If Part B does not cover the drugs, then Part D will cover them (if needed, original medicare beneficiaries can join a  Medicare drug plan).
However, if the patient only has Medicare due to End Stage Renal Disease, coverage will end 36 months after the kidney transplant.
In that case, and if the patient has no other health coverage, a special immunosuppressive benefit is available.
The patient will pay \$97.1 per month for this drug benefit; after meeting the deductible, the patient will pay 20\% of the Medicare-approved amount.
\end{scenario}

These health insurance scenarios were found in Medicare's official booklet on benefits \cite{knowyourmedicare}.
Scenarios 1 and 2 are comparable in verbosity and complexity, while scenario 3 is the most complex.
The boxes above contain paraphrased versions and the full policies can be found in our git repository~\cite{paper_repo}.

\subsection{Experiment Tasks}
We evaluate the following tasks as discussed in the Methodology section. We summarize these here for completeness:
\begin{itemize}
  \item \textbf{Task 1}. \textit{Write a structured text summary} including a concise summary and a bulleted list of concrete decisional criteria pertaining to health insurance processes.
  \item \textbf{Task 2}. \textit{Extract high-level domain logic} that encodes the process and decisional criteria from the natural text. We consider N3, a domain-agnostic rule language (Task 2.1), and CQL, a healthcare formalism (Task 2.2).
  \item \textbf{Task 3}. \textit{Generate unit tests} that describe the functions and expected outcomes of the Solidity smart contract and automate the testing. 
  \item \textbf{Task 4}. \textit{Generate smart contract code} that automates the natural text process on the Ethereum blockchain using Solidity.
\end{itemize}

\subsection{Experiment Metrics}
To assess the output of the LLM for tasks (1)-(4) for our experiment scenarios, we propose the following experiment metrics (with the relevant tasks in parenthesis):
\begin{enumerate}
    \item \textit{Completeness}: Whether each decisional criterion is covered ((1)-(4));
    \item \textit{Soundness}: Whether each decisional criterion is correctly encoded ((1)-(4));
    \item \textit{Clarity}: Whether the decisional criteria pertaining to health insurance processes are presented clearly ((1)) since our goal is supporting experts to extract domain logic and write code.
    \item \textit{Syntax}: Whether the syntax of the output is correct ((2), (3), (4));
    \item \textit{Functioning}: Whether the output code functions as intended ((2), (3), (4)).
\end{enumerate}

For checking \textit{syntax}, we use the appropriate compiler (N3, CQL, and Solidity).
Regarding the more subjective \textit{completeness}, \textit{soundness}, \textit{clarity}, and \textit{functioning} metrics, we employ the following strategy to mitigate the threat of subjective bias. 
We developed a scenario-specific rubric per metric, which grades the involved criteria based on their presence (completeness), soundness, clarity, and functioning. 
Subsequently, all 3 authors completed the rubrics, and inconsistencies were resolved through internal discussion.

\section{Experiment Results}
Table \ref{tab:summary} summarizes our experiment metrics for tasks (1)-(4) and scenarios (1)-(3).
In the result text, we annotate metric observations using a binary \cmark (positive) or \xmark (negative). For the same task, there may be multiple observations on the same metric; a negative may imply only minor issues.



\subsection{Comparison Between Models}
We find that the GPT-4 models perform better than the GPT-3.5 models; the latter performs similarly to the CodeLLaMA-34B model. 
It is worth noting that while CodeLLaMA is a smaller model, it performed better than our initial expectations and is an open-source model that can thus be freely experimented with. 
We mainly discuss the results from GPT-4 Turbo since it performed best---in our analysis below, we will refer to this model as \textit{the LLM}; 
when noteworthy, we outline differences between models.


\subsection{Task 1: Write a structured text summary.}
Compared to the other tasks, the LLM output is the most \textit{complete}, \textit{sound}, and \textit{clear} for this task, 
which is unsurprising as LLMs are adept at natural language output. 
The LLM does more than simply copy-pasting parts of the input. 
Overall, the high-level summary captures the requirements well, and 
the bulleted list does a good job of explicating the requirements found in the narrative input text.
That said, we point out some exceptions below:

\begin{itemize}
  \item \textit{Scenario 2}: 
  The LLM adds parts that were not strictly found in the input (\textit{soundness} \xmark):
  (1) IVIG is covered for patients ``who require treatment at home'', and that IVIG treatment ``can be administered at home.''
  The original text only mentions that ``it’s medically appropriate for you to get the IVIG in your home.''
  \item \textit{Scenario 3}:
  The textual summary does not explain all requirements found in the text (\textit{completeness} \xmark).
  The output mentions, ``[..] depends on various factors, such as having Medicare Part A and Part B at the appropriate times.''; however, the text describes when in particular Part A and Part B would be required.
  That said, this scenario has the highest complexity, and the prompt's task description indeed stipulates that \textit{``You must provide a high-level summary of the text}''.
  It may thus be argued that the response meets this goal.
  The bulleted list in the output explicated all requirements from the input text.
\end{itemize}

Among the tested models, the GPT-based models tend to break down the document in the most concise manner.
CodeLLaMA tended to group more requirements into a single bullet;
this behavior was noted throughout the scenarios regardless of their complexity.
We also note that the GPT-4 variants sectioned the bulleted list into multiple sub-lists, which was never specified in the input prompt. 
The sublists contained requirements related to the parent bullet, resulting in a more understandable summary.
Ultimately, we found that the summaries generated by GPT-4 models (both regular and turbo) were the easiest to follow.

\subsection{Task 2: Extract high-level domain logic.}

\subsubsection{Task 2.1 N3 Logic:}
We observe that the N3 syntax is mostly being adhered to for scenarios 1 and 2 (\textit{syntax} \cmark), with some minor exceptions.
After fixing these for scenarios 1 and 2, the N3 code functions as intended (\textit{functioning} \cmark).
No rules are generated for scenario 3 (see below).
Meaningful terms encode the decision logic (\textit{clarity} \cmark). 
We elaborate on \textit{soundness} and \textit{completeness} below.

A general observation across all three scenarios is that basic conceptual modeling principles are not being adhered to (\textit{soundness} \xmark).
For instance, the following triple was generated for scenario 1: 
\texttt{\small?fracture :certifiedBy ?doctor} and \texttt{\small?doctor :certifiesRelationTo :PostMenopausalOsteoporosis}; 
Here, we only know that a doctor certified a fracture, and that same doctor (possibly in this or another case) certified some relation with post-menopausal osteoporosis.
This is also referred to as a \emph{fan trap} in conceptual modeling~\cite{goelman2004entity}, and there are many more examples of these conceptual modeling issues.
Hence, when providing a full patient health record as input, the N3 code will likely infer false positives or negatives for this reason. 

Below, we comment on output for individual scenarios:

\noindent\textbf{\textit{Scenario 1}}: 
The LLM generated four rules in total, and there is one valid instance of rule chaining, i.e., a condition in a second rule referring to the inference or output of a first rule;
medicare covers the injection only if the patient meets the injection condition, as inferred by an earlier rule. 
This is a useful design pattern as it allows to modularize the decision logic into separate rules and makes the code clearer (\textit{clarity} \cmark).
That said, the first rule incorrectly infers for all patients that they have osteoporosis and meet the criteria for the home health benefit; the inference of the second rule is never referred to (both relate to \textit{soundness} \xmark).
Finally, the N3 rules miss the ``female'' criterion (\textit{completeness} \xmark).

\noindent\textbf{\textit{Scenario 2}}: One large rule is generated with all relevant conditions (clarity \xmark). 
Moreover, instead of inferring a final decision, its conclusion is quite literally taken from the prompt (\textit{soundness} \xmark): \texttt{\small?decision rdf:type :FinalDecision}, \texttt{\small?decision :fitsNaturalLanguageText true}. The prompt stipulated that ``Every logic rule must contribute to a final decision which fits the natural language text''; the conclusion should state that insurance will cover the IVIG.

\noindent\textbf{\textit{Scenario 3}}: 
No N3 rules are being generated at all; instead, reified statements (using RDF reification [X]) labeled \texttt{\small Rule [X]} are generated (\textit{soundness} \xmark).
Some of the more straightforward conditions are reflected in these reified statements (such as \emph{``part A coverage required-at transplant"}), however, 
more complex conditions are inserted as comments or string objects. 
In multiple cases, the reified part encodes what would be the ``conclusion'' part of the rule:
e.g., with comment \emph{``If Part B doesn’t cover immunosuppressive drugs,"} and the reified part encoding \emph{``part D covers immunosuppressive drugs.''}
However, all the comments and reified statements are clear (clarity \cmark) and seem to collectively represent all conditions (\textit{completeness} \cmark).
Interestingly, namespaces are used to add meaning to the terms, such as \texttt{\small partA:coverage} and \texttt{\small partB:requiredAt}, respectively indicating coverage of \emph{part A} and requirement for \emph{part B}.
Typically, namespaces provide specialized terms (e.g., predicates, classes) and do not indicate an object-property pair, as done here.

In general, we observe that scenario 3 exceeds the current ability of LLMs to generate code, as we make a similar observation for the smart contract output.

\subsubsection{Task 2.2 CQL:}
We observe that adherence to syntax is by far the most problematic for this target (\textit{syntax \xmark}), which we elaborate on below. Arguably, this is due to a lack of training data for CQL. There are ample code samples for N3 (having first been introduced in 2005, despite relatively low adoption) and Solidity (being a popular blockchain language, despite being only introduced in 2014). We are unable to locate the same amount of code samples for CQL on the Web; most examples are found in Github repos directly associated with CQL (such as \url{https://github.com/cqframework}).

In this vein, multiple essential syntactical elements were missing: 
(1) \textit{Code definitions}, which link decision logic to clinical terminologies (e.g., SNOMED-CT~\cite{donnelly2006snomed}, ICD-11~\cite{world1992icd}) for unambiguous clinical concepts; 
(2) \textit{Context statement} that scopes data that is accessible to the decision logic (e.g., context ``patient'' makes all patient-related data accessible).
It is difficult to comment on conceptual modeling, and \textit{soundness} in general, as the syntax does not allow an in-depth analysis.
That said, we observe a good attempt to modularize the code, 
with scenarios 1 and 3 featuring a ``final'' named expression referring to previous expressions that encode individual conditions.

Below, we comment on the individual scenarios to the extent possible:

\noindent\textbf{\textit{Scenario 1}}:
Named expressions, which are the building blocks of CQL libraries, are labeled incorrectly, i.e., using ``expression" instead of ``define," and incorrectly use the `\$' symbol (\textit{syntax} \xmark).
The named expressions further lack \textit{retrieve} and \textit{query} statements, essential elements of named expressions, which retrieve, constrain, and ``shape" the returned data (\textit{syntax} \xmark). 
Regarding \textit{completeness}, aside from the requirement for the doctor to certify the fracture cause, all requirements seem to be covered (\textit{completeness} \cmark).

\noindent\textbf{\textit{Scenario 2}}:
We found the syntax here to be more in line with what is expected (\textit{syntax} \cmark), with named expressions being labeled using `define' and including \textit{retrieve} and \textit{query} elements.
The \textit{retrieve} elements should refer to FHIR resources; a few of these resources exist (\textit{MedicationDispense(d)}, \textit{Coverage}), but most are being made up (e.g., \textit{MedicationActive}, \textit{CoveragePayer}) (\textit{functioning} \xmark).
As before, the code seems more or less complete (\textit{completeness} \cmark). However, this was difficult to ascertain due to syntax issues.

\noindent\textbf{\textit{Scenario 3}}:
Similar to scenario 2, we found the syntax here to be more in line with expectations (\textit{syntax} \cmark), with named expressions, retrieve and query elements.
However, the associated conditions are mostly incorrectly specified (\textit{functioning} \xmark).
Interestingly, the code mentions ICD10 codes \texttt{\small Z94.0} and \texttt{\small N18.6}, which indeed stand for kidney transplant and end-stage renal disease, respectively (both for reimbursement purposes).
On the other hand, code \texttt{\small 302803005} is mentioned, which is seemingly a SNOMED code, but it does not exist.
Several conditions, namely \texttt{\small inpatient} and \texttt{\small outpatient}, are being ``hallucinated''.
Finally, similar to before (Task 2.1), a long string is being inserted to explain a rather complex condition, i.e., the fact that a separate benefit may help if one loses 36 months after the organ transplant.

\begin{table*}[!htbp]
\centering
{\small
\begin{tabular}{|ll|l|l|l|l|l|}
\hline
\multicolumn{2}{|l|}{}                                                         & \textbf{Complete?}    & \textbf{Sound?}                & \textbf{Clear?}       & \textbf{Correct Syntax?} & \textbf{Functioning?}   \\ \noalign{\hrule height 1pt}
\multicolumn{1}{|l|}{\multirow{4}{*}{\textbf{Scenario 1}}} & \textbf{Task 1}   & Yes                   & Yes                            & Yes                   & N/A                      & N/A                     \\ \cline{2-7} 
\multicolumn{1}{|l|}{}                                     & \textbf{Task 2.1} & \textit{Minor issues} & \textit{Minor issues}\small{*} & Yes                   & \textit{Minor issues}    & Yes                     \\ \cline{2-7}
\multicolumn{1}{|l|}{}                                     & \textbf{Task 2.2} & \textit{Minor issues} & \textit{Minor issues}\small{*} & Yes                   & \textbf{Major issues}    & \textbf{\underline{No}} \\ \cline{2-7}
\multicolumn{1}{|l|}{}                                     & \textbf{Task 3}   & \textit{Minor issues} & Yes                            & \textit{Minor issues} & \textit{Minor Issues}    & Yes                     \\ \cline{2-7}
\multicolumn{1}{|l|}{}                                     & \textbf{Task 4}   & \textit{Minor issues} & \textit{Minor issues}\small{*} & \textit{Minor issues} & Yes                      & Yes                     \\ \noalign{\hrule height 1pt}
\multicolumn{1}{|l|}{\multirow{4}{*}{\textbf{Scenario 2}}} & \textbf{Task 1}   & Yes                   & \textit{Minor issues}          & Yes                   & N/A                      & N/A                     \\ \cline{2-7} 
\multicolumn{1}{|l|}{}                                     & \textbf{Task 2.1} & Yes                   & \textit{Minor issues}\small{*} & \textit{Minor issues} & \textit{Minor issues}    & Yes                     \\ \cline{2-7} 
\multicolumn{1}{|l|}{}                                     & \textbf{Task 2.2} & \textit{Minor issues} & \textit{Minor issues}\small{*} & Yes                   & \textbf{Major issues}    & \textbf{\underline{No}} \\ \cline{2-7}
\multicolumn{1}{|l|}{}                                     & \textbf{Task 3}   & Yes                   & Yes                   & \textit{Minor issues} & \textit{Minor issues}  & Yes                     \\ \cline{2-7}
\multicolumn{1}{|l|}{}                                     & \textbf{Task 4}   & Yes                   & Yes\small{*}                   & Yes                   & Yes                      & Yes                     \\ \noalign{\hrule height 1pt}
\multicolumn{1}{|l|}{\multirow{4}{*}{\textbf{Scenario 3}}} & \textbf{Task 1}   & \textit{Minor issues} & Yes                            & Yes                   & N/A                      & N/A                     \\ \cline{2-7} 
\multicolumn{1}{|l|}{}                                     & \textbf{Task 2.1} & Yes                   & \textbf{Major issues}\small{*} & Yes                   & \textbf{Major issues}    & \textbf{\underline{No}} \\ \cline{2-7} 
\multicolumn{1}{|l|}{}                                     & \textbf{Task 2.2} & \textit{Minor issues} & \textbf{Major issues}\small{*} & Yes                   & \textbf{Major issues}    & \textbf{\underline{No}} \\ \cline{2-7}
\multicolumn{1}{|l|}{}                                     & \textbf{Task 3}   & \textbf{Major issues} & \textbf{Major issues}         & \textbf{Major issues}                    & \textbf{Major issues}    & \textbf{\underline{No}}                      \\ \cline{2-7}
\multicolumn{1}{|l|}{}                                     & \textbf{Task 4}   & \textbf{Major issues} & \textbf{Major issues}*         & Yes                   & Yes                      & Yes                     \\ \hline
\end{tabular}
}
\caption{Summary of the Experiment Results from GPT-4 Turbo \emph{\small{(*: in addition to the conceptual modeling issues)}}}
\vspace{-4ex}
\label{tab:summary}
\end{table*}

\subsection{Task 3: Generate Unit Test Code}
Unit tests are used to check whether the target program functions as expected and to detect errors.
Thus, the unit test should not only check for correct outcomes but also cover every case of failure. 
Each unit test should be named clearly (\textit{clarity}) to describe which condition is being tested, cover every possible combination of inputs correctly (\textit{completeness} and \textit{soundness}), and be runnable (\textit{syntax} and \textit{functioning}).

The GPT-variant models use the correct syntax for Remix's testing plugin while CodeLLaMA sometimes diverges (\textit{syntax} \cmark). For example, we find instances where it uses \texttt{\small assert.equals} instead of \texttt{\small Assert.equal}.
We observe that the naming of the test cases is descriptive of the condition being tested (\textit{clarity} \cmark), but does not always imply the \textit{reason} the test should succeed or fail (\textit{clarity} \xmark).
For example, scenario 2 dictates that IVIG at home is covered if a patient has a \emph{``diagnosis of primary immune deficiency disease''}.
The LLM generated a corresponding test named \texttt{\small testIVIGCoverageNotCovered},
but this name does not cover the reason (i.e., the aforementioned diagnosis).

%
We note that the LLM tends to only generate tests for conditions that are explicitly formulated as such in the text (\textit{completeness} \xmark).
For example, in scenario 1, \texttt{\small home health nurse} should not be covered if the patient is not diagnosed with Osteoporosis, but the model does not generate a test to check whether the patient has a disease at all.
Such requirements may be obvious to the human expert, but it appears to be something LLMs tend to overlook.
%
We describe the output of the LLM for each scenario below:

\noindent\textbf{\textit{Scenario 1}}:
The LLM generates a set of tests that cover the \texttt{\small injectable osteoporosis drug} condition from the document, but the \texttt{\small home health nurse} condition is not always covered (\textit{completeness} \xmark).
The tests correctly describe the conditions that should be met for the patient to be eligible for coverage (\textit{soundness} \cmark).


\noindent\textbf{\textit{Scenario 2}}:
The unit tests cover the necessary conditions described in the document (\textit{completeness} \cmark).
We also note that the model generates a set of instances of the \textit{struct} (an abstract data type representation in Solidity) with different values for the properties, which is a good practice for unit testing and will support more use cases (\textit{completeness} \cmark).

\noindent\textbf{\textit{Scenario 3}}:
As expected, the LLM struggles the most with this scenario with several missing conditions (\textit{completeness} \xmark).
None of the outputs use a struct to define the contract being tested and sometimes does not explicitly state which inputs are mapped to which value (\textit{clarity} \xmark).
For instance, instead of using variables such as \texttt{\small isPartACovered = true; isCovered(isPartACovered, ...)}, the test checks for \texttt{\small isCovered(true, ...)}, which results in incomprehensible code.
We also find instances where multiple cases are handled in one test, which is bad practice for unit tests (\textit{clarity} \xmark). 
It appears that the LLM struggles with the length of the input document.
When the input document is large, the models show a tendency to group multiple conditions into one case, even when the conditions are not necessarily related to each other.
For example, the document mentions that a benefit provided by Medicare is not \emph{``a 
substitute for full health coverage,''}, which is not a testable condition and is not related to the coverage at hand.
However, we find that the LLM attempts to test this statement in its unit tests (\textit{soundness} \xmark).
We also find trivial issues with units, such as using ``month'', which is incorrect Solidity syntax, or ``ether'' for a dollar amount (\textit{syntax} \xmark).

\subsection{Task 4: Generate Smart Contract Code}
We observe that the Solidity syntax is being adhered to, i.e., no syntax errors were found for scenarios 1 and 2 (\textit{syntax} \cmark),
and the smart contract functions as intended (\textit{functioning} \cmark), with the same caveat as N3 on conceptual modeling.

Similar to the previous tasks, we found that meaningful attributes and method names are utilized (\textit{clarity} \cmark).
However, basic conceptual modeling principles are not being adhered to (\textit{soundness} \xmark).
For instance, the following property is used to check whether the patient has a bone fracture related to post-menopausal osteoporosis: \texttt{\small patient.hasFracture \&\& patient.isRelatedToOsteoporosis}, which checks whether the patient has a fracture and whether any aspect related to the patient is related to osteoporosis, failing to confirm the relation between the two.
There are many more examples of these issues, which means that when providing a full patient health record, the smart contract may issue incorrect results (\textit{soundness} \xmark).
In line with this observation, none of the smart contracts include more than one struct, although all scenarios include multiple concepts, such as \emph{patients, doctor, insurance coverage (and different parts), transplants}, which could easily be modelled in different structs (clarity \xmark).

Below, we comment on output for individual scenarios:

\noindent\textbf{\textit{Scenario 1}}:
The LLM generates one large method containing all relevant conditions (we note that the N3 code was better modularized) (\textit{clarity} \xmark).
Also, there are some issues with completeness and soundness. Examples include: the generated code lacked a required relation between bone fracture and post-menopausal osteoporosis (\textit{completeness \xmark}), and the coverage of a home health nurse is a separate reimbursement but is evaluated within the same main reimbursement checking method (\textit{soundness \xmark}).

\noindent\textbf{\textit{Scenario 2}}:
The LLM generates one large method with only three conditions (acceptable \textit{clarity} \cmark), but it is otherwise sound (\textit{soundness \cmark}) and complete (\textit{completeness \cmark}).

\noindent\textbf{\textit{Scenario 3}}:
The LLM generates multiple methods with meaningful names that contain the relevant conditions (\textit{clarity} \cmark). 
However, many conditions from the input text are missing, e.g., \emph{require part B at the time of drug prescription, alternative coverage by part D, ESRD, and 36-month time limit}, etc (\textit{completeness \xmark}).
We find a similar issue with the N3 code, as the scenario seems too complex for the LLM to generate appropriate smart contract code.
Most properties involved in these conditions are found in the struct but are not referenced in the methods.
Hence, there is a distinct lack of \textit{completeness} as noted above, in addition to \textit{soundness} (\textit{soundness} \xmark).
As with  N3, many comments are seemingly generated to cope with the added complexity. For example, we found the following at the bottom of the output from the LLM: ``Please note that this is a simplified implementation that focuses on the logic described in the text.''
The code attempts to (incorrectly) calculate the total patient cost, but this seems to feature a ``hallucination'', as it refers to co-insurance, which is not found in the text (\textit{soundness} \xmark).
Finally, as a detail, a series of \texttt{\small require} statements are used to check whether all conditions are met (throwing an error if the passed value is false); the other scenarios included if/then statements.
So, it appears that the LLMs defaulted to better software engineering in this case (clarity \cmark).

\section{Related Work}
Automated code generation has a rich research history spanning various domains, extending beyond the specific focus on health insurance in this work. The advent of LLMs has catalyzed a surge in studies centered around LLM-based code generation. Among these, Codex, which is a GPT language model refined through fine-tuning on publicly available GitHub code, stands out prominently and represents a significant development in the field~\cite{chen2021evaluating}.
\citet{vaithilingam2022expectation} presents a comprehensive user study focusing on the usability of GitHub Copilot that uses Codex. They observe that while advances in deep learning have shown promise in automatic code generation, it often suffers from low accuracy and reliability.
\citet{ni2023lever} proposes an approach, called LEVER to enhance language-to-code generation by verifying the generated programs based on their execution results. This process involves re-ranking the sampled programs from LLMs, considering a verification score combined with the LLM generation probability. 
While LEVER demonstrated significant performance improvements over baseline methods, \citet{ni2023lever} acknowledges limitations like the dependence of program execution on specific inputs and contexts, where some edge cases were not handled properly.
To improve the accuracy of generated code, in future work, we describe the possibility of manually fixing intermediary output and providing this as partial input to subsequent steps.

Specific to healthcare, \citet{khanChatGPTReshapingMedical2023} discuss possible uses of LLMs in clinical management, such as documentation or generation of clinical decision support systems.
They also note that while the output of the models is impressive, it should not be used as a replacement for human agents. 
\citet{benoitChatGPTClinicalVignette2023} present their results from testing OpenAI's ChatGPT's ability to handle medical information.
The model was asked to provide vignettes about a pediatric patient's status, assuming different levels of medical literacy, to test its knowledge of medical terminology.
The model was then asked to generate a vignette and a diagnosis from a clinician's point of view based on symptoms extracted from existing medical documents, and its output was compared to a real vignette made by a doctor. 
\citet{benoitChatGPTClinicalVignette2023} note that while ChatGPT demonstrates an exceptional usage of medical vocabulary, its output is unreliable and must be monitored carefully. 
\citet{jeblickChatGPTMakesMedicine2022} presents a study in which 15 radiologists assessed the output of ChatGPT's simplified reports.
While most radiologists reported that ChatGPT's summary was truthful and not harmful to the patient in simple cases, they also noted cases that overlooked key aspects and outputted potentially harmful information. 
Contrary to these related works, we focus on a niche application that applies LLMs on textual health insurance policies, which combine legal, financial and health-related knowledge, in order to automate processes on blockchain. 
To the best of our knowledge, this is an area that has not been explored previously.

\section{Conclusion}
We present an initial exploration of using LLMs for generating decision logic and smart contracts for healthcare insurance processes from natural text, applied to multiple real-world cases from the official medicare coverage guidebook \cite{knowyourmedicare}. 
The experimental results show that while state-of-the-art LLMs exhibit remarkable performance in generating structured text summaries, human oversight is crucial when automating the more complex tasks of generating formal logic and executable smart contracts. 
We observe that more complex scenarios (\textit{scenario 3}) exceed the current ability of LLMs to generate corresponding code, and formalisms with less online training data, such as CQL, are not suitable targets for code generation.
Additionally, we note a deviation from the core principles of conceptual modeling, which compromises the \textit{soundness} of the output when providing a full patient record.
The model sometimes overlooks important details, such as missing unit test conditions (e.g., having Osteoporosis) that are mentioned in the text but not formulated as if-then conditions.
We also notice instances of \textit{hallucinations} in the model outputs, where conditions not mentioned in the document appear in the encoded output. 
This phenomenon also occurs more frequently when the document is lengthy or complex.

Despite these limitations, our experiments highlight the promise of LLMs in automating the generation of smart contracts for health insurance policies.
%
Improvements can be made to LLMs to consider basic conceptual modeling principles and enhance the generated coding style.
Incorporating coding conventions, industry standards, and conceptual modeling principles into the training process of the LLM can help produce more sound and clear decision logic and smart contract code.
This can include techniques such as fine-tuning the model on coding style and conceptual modeling  guidelines and integrating associated quality metrics into the evaluation process.
Furthermore, complex healthcare scenarios often involve intricate logic and conditional statements.
Future work can focus on enhancing the LLM's ability to capture and represent such complexity accurately.
This can involve training the model on diverse and complex documents and incorporating more advanced language understanding techniques, such as multi-hop reasoning and knowledge graph integration. 

We acknowledge the possibility that different prompts may yield different results. Therefore, to assess the robustness of our findings, we will explore prompt variations to determine the sensitivity of the LLM's output to changes in prompt wording and structure.
Moreover, we will experiment with alternative prompt engineering approaches. 
\cite{weiChainofThoughtPromptingElicits2022} report that using a series of related prompts instead of a single prompt each time leads the LLM to perform better on the downstream task.
The authors call this approach \textit{chain-of-thought}.
Inspired by this approach, we will experiment with multiple code generation pipelines:
\begin{itemize}
    \item Send the original natural text and the output of the prior task for every task.
    \item Send the entire conversation history of the task for every task---i.e., the original natural text and output of every previous task. 
\end{itemize}

By embracing advancements in LLM technology, 
we can anticipate enhanced capabilities and increased efficiency in automating processes in healthcare and other domains reliant on dense textual information. 
For instance, this line of work will open up possibilities for process automation in various healthcare settings, including medical billing, patient record management, and clinical research.
However, healthcare regulations, ethical considerations, and domain-specific requirements (rightly) still necessitate the involvement of domain experts and legal professionals to ensure compliance and mitigate risks. 

The full prompts, experiment text of the scenarios, and experiment results can be found in our GitHub repository~\cite{paper_repo}.

\section*{Acknowledgements}
The authors acknowledge the support from NSF IUCRC CRAFT center research grant (CRAFT Grant \#22008) for this research. The opinions expressed in this publication do not necessarily represent the views of NSF IUCRC CRAFT.

\bibliographystyle{spbasic}
\bibliography{references}

\end{document}